\documentclass[sigconf,screen]{acmart}

\AtBeginDocument{%
  }

\usepackage{times}

\usepackage{soul}
\usepackage{url}
\usepackage{hyperref}
\usepackage[utf8]{inputenc}
\usepackage{caption}
\usepackage{graphicx}
\usepackage{amsmath}
\usepackage{booktabs}
\usepackage{tabularx}
\usepackage[ruled,vlined]{algorithm2e}
\usepackage{multirow}
\usepackage{algorithm2e}
\usepackage{tikz}
\usetikzlibrary{shapes, arrows.meta, positioning, calc, fit, backgrounds}

\newcounter{observation}
\newcommand{\observation}[1]{\refstepcounter{observation}
    \vspace{0.1cm}
    \noindent\fbox{\parbox{0.95\columnwidth}{\textbf{Observation \theobservation:} #1}}
    \vspace{0.1cm}
}

\setcopyright{none}
\copyrightyear{}
\acmYear{}
\acmDOI{}
\acmISBN{}
\acmConference{}
\acmBooktitle{}
\acmPrice{}

\begin{document}

\title{AST-PAC: AST-guided Membership Inference for Code}





\author{Roham Koohestani, Ali Al-Kaswan, Jonathan Katzy, Maliheh Izadi}
\affiliation{%
  \institution{Delft University of Technology}
  \city{Delft}
  \country{Netherlands}}
\email{{RKoohestani, A.Al-Kaswan, J.B.Katzy, M.Izadi}@tudelft.nl}

\renewcommand{\shortauthors}{Koohestani et al.}

\begin{abstract}
Code Large Language Models are frequently trained on massive datasets containing restrictively licensed source code. This creates urgent data governance and copyright challenges. Membership Inference Attacks (MIAs) can serve as an auditing mechanism to detect unauthorized data usage in models. While attacks like the Loss Attack provide a baseline, more involved methods like Polarized Augment Calibration (PAC) remain underexplored in the code domain. This paper presents an exploratory study evaluating these methods on 3B--7B parameter code models. We find that while PAC generally outperforms the Loss baseline, its effectiveness relies on augmentation strategies that disregard the rigid syntax of code, leading to performance degradation on larger, complex files. To address this, we introduce \textbf{AST-PAC}, a domain-specific adaptation that utilizes Abstract Syntax Tree (AST) based perturbations to generate syntactically valid calibration samples. Preliminary results indicate that AST-PAC improves as syntactic size grows, where PAC degrades, but
under-mutates small files and underperforms on alphanumeric-rich code.
Overall, the findings motivate future work on syntax-aware and size-adaptive calibration as a prerequisite for reliable provenance auditing of code language models.
\end{abstract}



\maketitle

\section{Introduction}
\label{sec:intro}
Large Language Models (LLMs) are trained on web-scale data~\cite{brown2020language}, often compiled with minimal oversight regarding personal or copyrighted content~\cite{katzy2024exploratory, alkaswan2023abuse}. This has sparked legal challenges regarding \textit{fair use} and copyright infringement~\cite{NYT_2024_Lawsuit,RopesGray_2025_FairUse, samuelson2025does}. In software engineering, this risk is amplified~\cite{yang2024unveiling, salerno2025much}; code LLMs have been shown to memorize license information verbatim, introducing risks of license infringement~\cite{al2024traces}.

To mitigate these risks, organizations require internal auditing mechanisms to verify whether specific content was consumed during training~\cite{yang2024unveiling, al2024traces}. Membership Inference Attacks (MIAs) are widely used to measure such data leakage~\cite{hu2022membership, wu2025membership, niu2024survey, yang2024robustness, meeus2025sok, wan2024does, zhang2024code}. In a grey-box auditing setting, where the auditor has access to model parameters, the choice of attack signal is critical. While the \textit{Loss Attack}~\cite{shokri2017membership, carlini2022membership} is computationally efficient, newer methods like Polarized Augment Calibration (PAC)~\cite{ye2024data} utilize more complex calibration signals. However, recent surveys note a lack of systematic evaluation of these methods on large-scale models~\cite{wu2025membership}, specifically regarding source code.

This paper presents an exploratory study addressing three research questions:
\begin{itemize}
    \item \textbf{RQ1 (Performance):} How does PAC compare to the Loss attack on code-specific LLMs?
    \item \textbf{RQ2 (Robustness):} How do data characteristics (e.g., complexity, size) affect attack success?
    \item \textbf{RQ3 (Adaptation):} Can a code-specific variant of PAC (AST-PAC) improve performance?
\end{itemize}

We evaluate MIAs as a technical mechanism for auditing training-data provenance
in code models and contribute: (i) a baseline comparison of Loss vs.\ PAC on
3B--7B code models, (ii) a stratified robustness analysis over file properties,
and (iii) \textbf{AST-PAC}, a syntax-aware calibration variant that replaces
token swaps with Tree-Sitter guided perturbations. The code and results for our study can be found in the replication package.\footnote{\href{https://drive.google.com/file/d/1P8CRpjThLAcriLTHdN4TmoOK7vm-CAmD/view?usp=sharing}{https://drive.google.com/file/d/1P8CRpjThLAcriLTHdN4TmoOK7vm-CAmD/view?usp=sharing}}

\section{Background and Methodology}
\label{sec:bg}
We consider a grey-box auditing setting where the auditor can compute token-level
log-probabilities (logits).

\subsection{Attack Signals}
\noindent\textbf{Loss Attack.} The baseline method~\cite{yeom2018privacy, carlini2021extracting} assumes member samples yield lower loss than non-members. For a sequence $x$, the score is $s_{\text{loss}}(x) = -\ell(x)$, where $\ell(x)$ is the negative log-likelihood (token cross-entropy).

\textbf{Polarized Augment Calibration (PAC).}~\cite{ye2024data} uses a reference-free \textit{polarized distance} ($L_M$) measuring the gap between the most and least confident tokens. It calibrates this score using \textit{adjacent samples} generated by randomly swapping tokens. The score definition is:
\[
s_{\text{pac}}(x) = L_M(x) - \frac{1}{n} \sum_{i=1}^{n} L_M(\tilde{x}_i)
\]
where $\tilde{x}_i$ are the augmented neighbors and $n$ is the count of those neighbors. A limitation of PAC for code is that random token swaps break syntax, which has been shown to disproportionally increase loss for code models.

\subsection{Experimental Setup}
\textbf{Data:} We use the Java subset of The Heap~\cite{katzy2025heap}, stratified into \emph{members}, \emph{near-members} (near-duplicates), and \emph{non-members}. In this context, near-members are samples
with Jaccard similarity $\geq 0.7$ to a member datapoint when using Min-Hash Locality Sensitive Hashing. For the details, we refer the reader to the original dataset paper~\cite{katzy2025heap}. 

\textbf{Models:} We evaluate four 3--7 billion parameter models: \textbf{Mellum-4B~\cite{pavlichenko2025mellum}} (code-completion), \textbf{StarCoder2-3B and StarCoder2-7B~\cite{lozhkov2024starcoder}} (code-specialized), and \textbf{SmolLM3-3B~\cite{bakouch2025smollm3}} (general-purpose) which we use to control whether general-purpose training dilutes the signal achievable through the attacks.
\textbf{Protocol:} We report the Area Under the Curve for the Receiver Operating Characteristic (ROC-AUC) and the Precision Recall curve (PR-AUC). We report ROC-AUC for threshold-independent discriminability and PR-AUC because it is sensitive to performance under class imbalance.
Attacks use default PAC parameters ($k_{near}=30, k_{far}=5, m=0.3, n=5$)~\cite{ye2024data}. 

\section{RQ1: Comparative Performance}
\label{sec:rq1}

We first establish a baseline comparison between PAC and Loss attacks.
\begin{table}[tb]
    \centering
    \caption{Aggregate Attack Performance (ROC-AUC / PR-AUC). Best scores per model are \textbf{bold}.}
    \label{tab:rq1}
    \resizebox{\columnwidth}{!}{
    \begin{tabular}{llcc}
    \toprule
    \textbf{Model} & \textbf{Scenario} & \textbf{Loss Attack} & \textbf{PAC Attack} \\
    \midrule
    \multirow{3}{*}{Mellum} 
        & All vs NM & 0.628 / 0.764 & 0.696 / \textbf{0.827} \\
        & Exact vs NM & 0.629 / 0.628 & \textbf{0.712} / 0.742 \\
        & Near vs NM & 0.626 / 0.617 & 0.679 / 0.691 \\
    \midrule
    \multirow{3}{*}{SmolLM} 
        & All vs NM & 0.573 / 0.701 & 0.578 / \textbf{0.735} \\
        & Exact vs NM & 0.572 / 0.539 & 0.579 / 0.591 \\
        & Near vs NM & 0.575 / 0.541 & \textbf{0.578} / 0.579 \\
    \midrule
    \multirow{3}{*}{SC2 3B} 
        & All vs NM & 0.625 / 0.750 & \textbf{0.662 / 0.798} \\
        & Exact vs NM & 0.622 / 0.600 & \textbf{0.662} / 0.680 \\
        & Near vs NM & 0.629 / 0.605 & 0.661 / 0.664 \\
        \midrule
    \multirow{3}{*}{SC2 7B} 
        & All vs NM & 0.635 / 0.758 & 0.675 / \textbf{0.806} \\
        & Exact vs NM & 0.634 / 0.611 & \textbf{0.678} / 0.694 \\
        & Near vs NM & 0.637 / 0.614 & 0.672 / 6.74 \\
    \bottomrule
    \end{tabular}
    }
\end{table}
\autoref{tab:rq1} shows that PAC improves over Loss on code-specialized models
(Mellum, StarCoder2), while gains are smaller on the general-purpose SmolLM3.
Near-members are detected nearly as effectively as exact members, suggesting
that minor edits do not reliably remove membership signal.

\observation{PAC provides a stronger, albeit rather low, auditing signal than raw Loss for code models (StarCoder2/Mellum). Both methods are robust to minor code variations (Near-Members).}

\begin{figure*}[tb]
    \centering
    \includegraphics[width=0.85\linewidth]{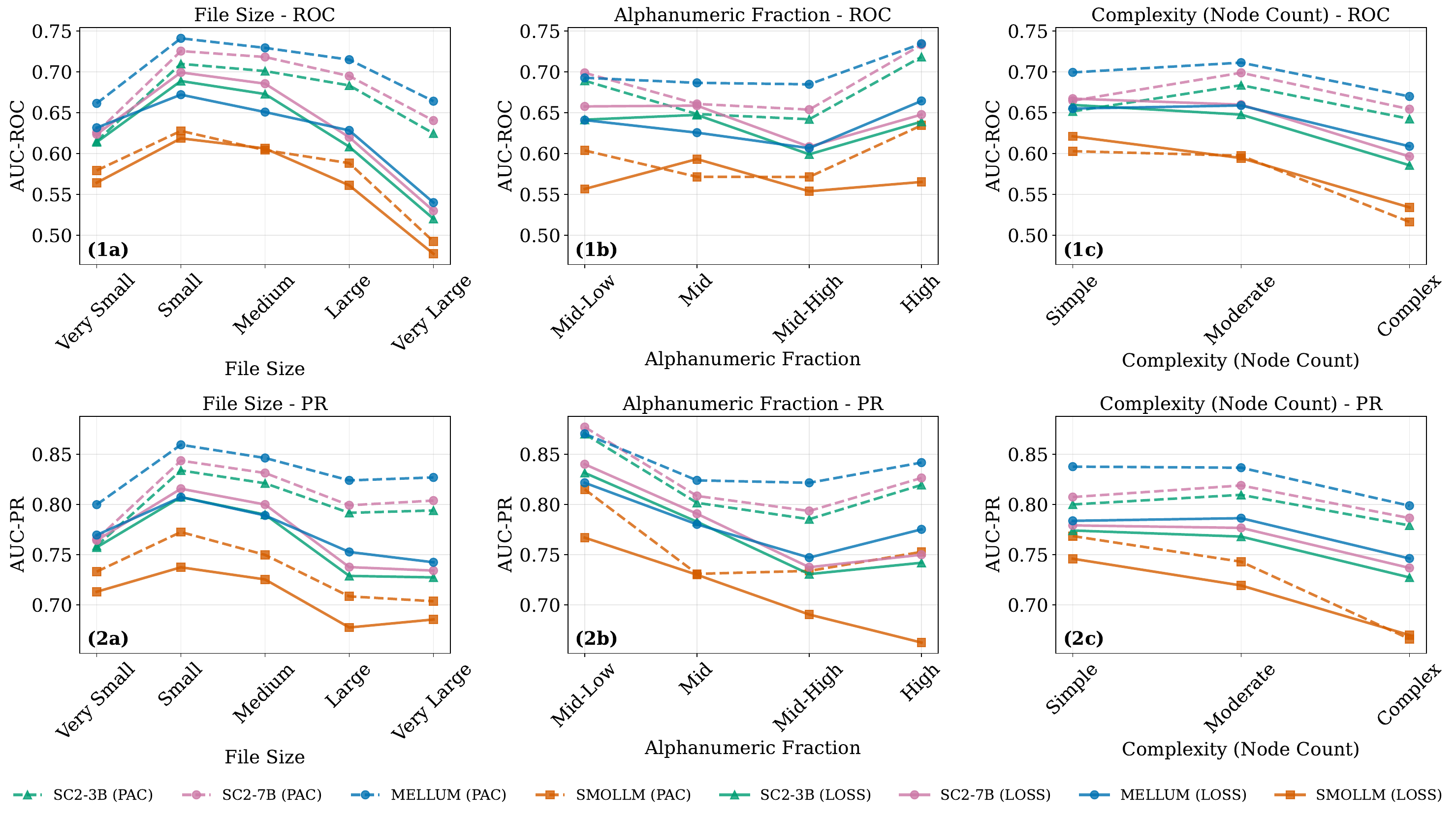}
    \caption{Experimental results for models and the loss and pac attacks grouped into categories based on different stratification of code attributes. Results for ROC-AUC (top row), and PR-AUC (bottom row).}
    \label{fig:rq2:fig}
\end{figure*}

\section{RQ2: Impact of Data Characteristics}
\label{sec:rq2}

\begin{table}[tb]
\centering
\caption{Stratification bucket cutoffs used in RQ2/RQ3.}
\label{tab:buckets}
\setlength{\tabcolsep}{4pt}
\begin{tabular}{lcc}
\toprule
\textbf{Dimension} & \textbf{Buckets} & \textbf{Numeric cutoffs} \\
\midrule
File size          & VS/S/M/L/VL & 1074, 1946, 3369, 6844 \\
AST nodes          & Simple/Mod/Comp & 58, 211 \\
Alphanum. ratio    & ML/M/MH/H & 0.2, 0.4, 0.6, 0.8 \\
\bottomrule
\end{tabular}
\end{table}

We apply stratified sampling to understand how code properties affect vulnerability. We analyze file size, alphanumeric ratio (density of identifiers/comments), and syntactic complexity (AST node count).
\autoref{tab:buckets} defines the numeric ranges underlying each bucket label.
Results (summarized in \autoref{fig:rq2:fig}) reveal distinct trends:
\begin{enumerate}
    \item \textbf{File Size:} An inverted U-shape is observed. Attacks peak at Small/Medium sizes but degrade on Very Large files. However, PAC is more robust to this degradation (StarCoder2 ROC-AUC 0.630 vs Loss 0.522 on Large files).
    \item \textbf{Alphanumeric Ratio:} Code with higher ratios (more natural language) is significantly easier to detect. On StarCoder2, PAC jumps from 0.641 (Mid-High) to 0.721 (High).
    \item \textbf{Complexity:} A negative correlation exists between AST node count and detection. Simple structures are easier to infer than complex ones.
\end{enumerate}

\observation{Attack effectiveness is highly data dependent. Files with high alphanumeric ratios
and files that are small-to-medium length provide the most reliable membership signal. very very large and complex files reduce attack reliability. 
PAC is more robust than Loss in large-file regimes.}

\section{RQ3: Code-Specific Adaptation}
\label{sec:rq3}

Standard PAC augmentation disregards syntactic and structural correctness. We hypothesize that this contributes to the performance drop on longer files:
token swaps increasingly disrupt syntactic structure and yield neighbors that
are poor ``adjacent'' samples for calibration.
To mitigate this, we introduce \textbf{AST-PAC}.

\SetKwProg{Fn}{Function}{:}{end}

\subsection{Method: AST-PAC}
\begin{algorithm}[h]
\caption{AST-PAC Neighbor Generation}
\label{alg:ast_pac}
\SetAlgoLined
\KwIn{Code sample $x$, mutation ratio $m$, max tries $T$}
\KwOut{Perturbed Sample $\tilde{x}$}

\vspace{0.2em}
\Fn{\textsc{GenerateNeighbor}$(x, m, T)$}{
  \For{$t \leftarrow 1$ \KwTo $T$}{
    $\mathcal{T} \leftarrow \textsc{Parse}(x)$\;
    $\mathcal{C} \leftarrow \textsc{CollectNodesByCategory}(\mathcal{T})$\;
    $\mathcal{C}' \leftarrow \{c \in \mathcal{C} \mid |c| \ge 2\}$\;
    \If{$\mathcal{C}' = \emptyset$}{\Return $x$}

    $N \leftarrow \sum_{c \in \mathcal{C}'} |c|$\;
    $B \leftarrow \max(1, \lfloor m \cdot N / 2 \rfloor)$\;
    $\{B_c\}_{c \in \mathcal{C}'} \leftarrow \textsc{AllocateBudgets}(B, \{|c|\})$\;

    $x' \leftarrow x$\;
    \ForEach{$c \in \mathcal{C}'$}{
      Re-parse $x'$ to refresh node offsets\;
      Select $2B_c$ nodes from $c$ without replacement\;
      $x' \leftarrow \textsc{SwapNodes}(x', \text{selected nodes})$\;
    }
    \If{$x' \ne x$}{\Return $x'$}
  }
}
\end{algorithm}

AST-PAC replaces token-swap augmentation in PAC with AST-guided
\emph{span permutations} over syntactically meaningful node categories (e.g.,
variable declarators, literals, top-level statements). For each input $x$, we
generate $n$ neighbors $\tilde{x}_i$ by selecting pairs of the same category and swapping their contents while preserving the structure of their surroundings
(\autoref{alg:ast_pac}). We then compute the calibrated score using the same
polarized-distance signal as PAC.
This targets syntactically plausible neighbors to reduce calibration noise
caused by invalid token swaps in code.

\subsection{Results}

\begin{figure*}[tb]
    \centering
    \includegraphics[width=0.85\linewidth]{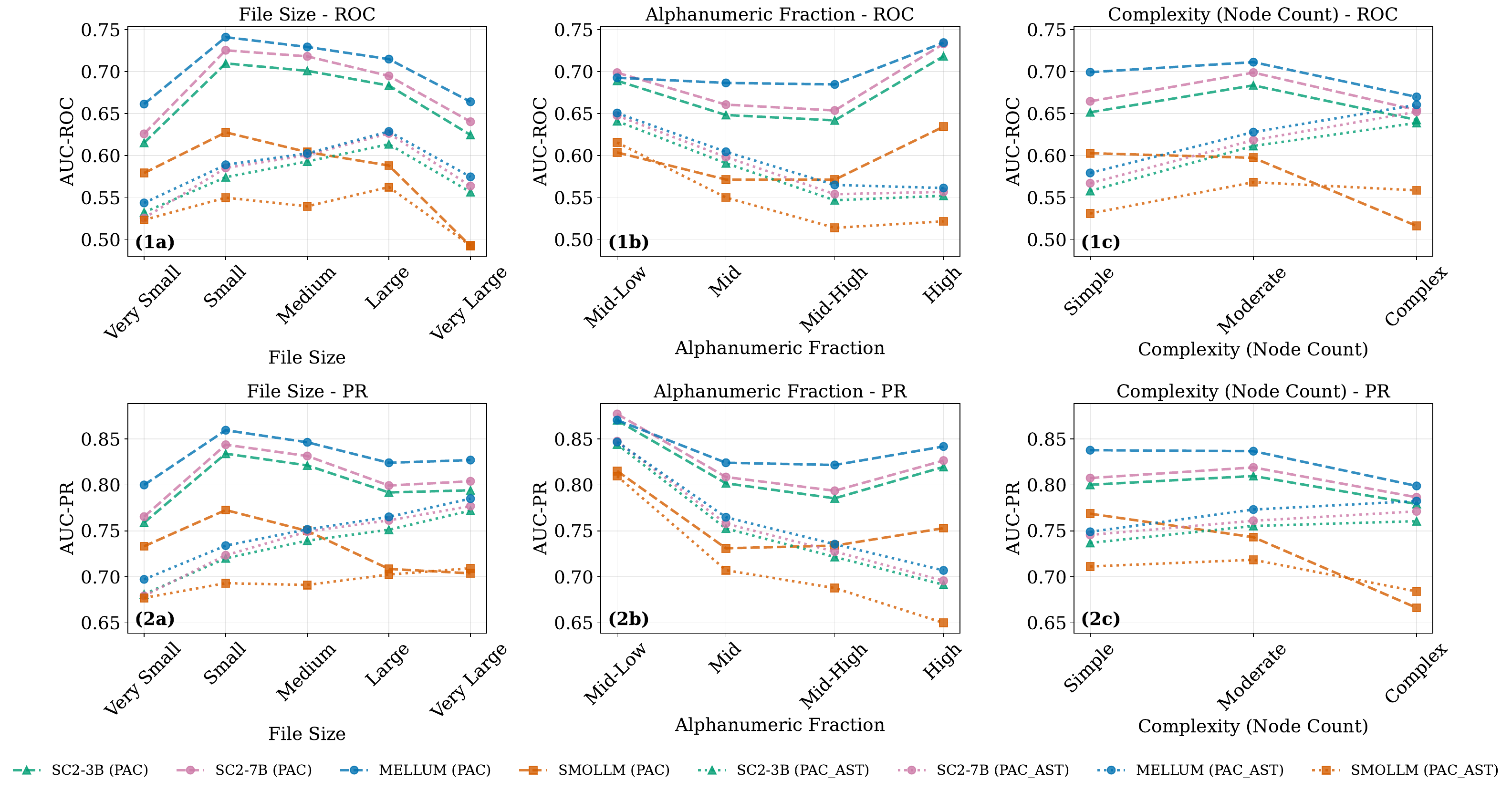}
    \caption{Experimental results for models and the pac and pac\_ast attacks grouped into categories based on different stratification of code attributes. Results for ROC-AUC (top row), and PR-AUC (bottom row).}
    \label{fig:rq3:pacvast}
\end{figure*}

\begin{figure}[tb]
  \centering
  \includegraphics[width=\columnwidth]{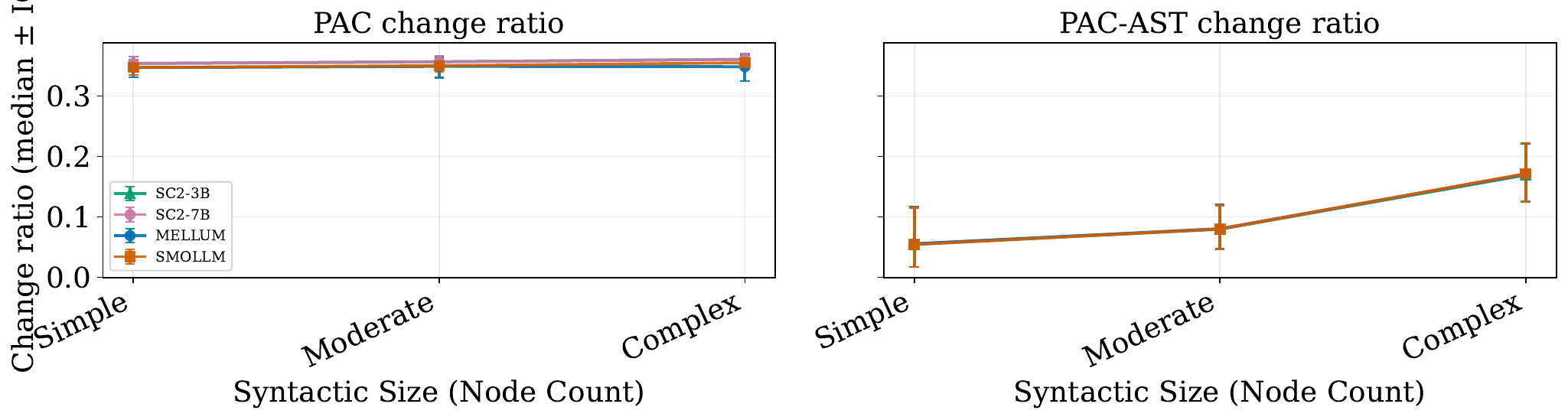}
  \caption{Effective change ratio by syntactic size.}
  \label{fig:rq3:change}
\end{figure}

AST-PAC changes the neighbor distribution used for calibration. Although both
methods use the same nominal mutation ratio ($m=0.3$), the \emph{effective}
change ratio differs; token-swap PAC remains near-constant, while AST-PAC yields
substantially smaller changes on files with fewer AST nodes
and increases with
node count (\autoref{fig:rq3:change}). 

Across code-specialized models, AST-PAC improves as file and syntactic size
increase: ROC-AUC and PR-AUC tend to rise from small/simple to large/complex
buckets, whereas PAC degrades in the large-file regime. In contrast,
AST-PAC performs worse on alphanumeric-rich code, where token swaps still yield
strong calibration signal, indicating that the current AST categories do not
perturb lexical cues (identifiers/comments) at comparable rates.

\observation{AST-PAC improves robustness in large/complex files where token-swap
PAC degrades, but it under-mutates small files and underperforms on
alphanumeric-rich code, motivating adaptive mutation budgets and hybrid
lexical+syntactic perturbations.}

\section{Discussion}
\label{sec:discussion}
Our results highlight distinct behaviors in how code models reveal their training data. We structure our discussion around four key observations regarding model specialization, robustness to modification, and data characteristics.

\textbf{Vulnerability of Specialized Models.}
We found that specialized code models (StarCoder2, Mellum) are significantly more susceptible to membership inference than the general-purpose control (SmolLM3). This suggests that focused training procedures lower the threshold for memorization, particularly for high-frequency structures like license text~\cite{al2024traces}. For organizations, this implies that specialized models carry a heightened audit risk and require more rigorous governance than general LLMs.

\textbf{Provenance Survives Refactoring.}
Our experiments show that Near-Members are detected with accuracy virtually identical to Exact Members (e.g., StarCoder2 PAC ROC-AUC: 0.663 vs 0.664). This indicates that models memorize underlying semantic structures or unique syntactic patterns rather than just verbatim token sequences. For auditors, this validates MIAs as robust tools for detecting infringement, as simple code refactoring is insufficient to erase the training signal.

\textbf{The ``Sweet Spot'' of Memorization.}
We identified that files with a \textbf{High Alphanumeric Ratio} (rich in natural language, comments, and identifiers) and \textbf{Small-to-Medium size} represent the most vulnerable stratum. 
LLMs seem to grasp the natural-language-like, high-entropy aspects of code more easily than its purely abstract logical structure.
Performance degrades on very large files due to possible loss signal dilution, though PAC proves more robust in these edge cases than Loss.

\textbf{When syntax-aware calibration helps.}
AST-PAC suggests that the calibration distribution, not only the underlying
signal, is a primary driver of robustness. Syntax-aware perturbations improve
separability in large and structurally complex files, where token swaps yield
increasingly unrealistic neighbors. However, AST-PAC under-mutates small files
and is weaker on alphanumeric-rich code, indicating that effective perturbation
magnitude and lexical coverage (on text, similar to PAC) must be controlled to obtain consistent gains.

\subsection{Threats to Validity}
\label{sec:discussion:threats}
\textbf{Metric Overlap:} AST node count correlates with file size; thus, effects attributed to complexity may partially reflect sequence length. 
\textbf{Model Scale:} We limited our study to 3-7B parameters due to resource constraints; larger models generally exhibit higher memorization capacities. 
\textbf{Language Bias:} Our focus on Java may favor AST-based methods due to its rigid syntax; results may differ for dynamic languages like Python.
\textbf{Near-member label ambiguity:} We evaluate near-members as members as a probe for robustness to minor edits. However, near-duplicate similarity does not necessarily reflect training membership. High-frequency constructs and boilerplate, including license templates, could lead to high similarity even without files being observed during training. This introduces label noise in the near-member setting.
\textbf{Residual contamination and label noise:} ``Non-member'' is defined
relative to the evaluation split, not the model's complete (partly unknown)
pretraining mixture. For broadly trained models (e.g., SmolLM3), some samples
labeled non-member may still have been seen during pretraining, reducing measured
separability and complicating cross-model comparisons.
\textbf{Correlated samples:} near-duplicates and repository-level similarities
can violate independence assumptions and make uncertainty estimates optimistic.
\textbf{Non-equivalent perturbation magnitude:} AST-PAC and token-swap PAC share
the same nominal $m$ but induce different effective change ratios across file
sizes, meaning the augmentation mechanism can confound the perturbation magnitude.

\section{Conclusion}
\label{sec:conclusion}
We evaluated grey-box membership inference on code LLMs and found that PAC
provides a stronger auditing signal than Loss on code-specialized models, while
near-duplicates remain detectable at similar rates to exact members. Attack
success depends strongly on program characteristics and degrades on very large
and structurally complex files. To mitigate syntactically implausible token-swap
neighbors, we introduced \textbf{AST-PAC}, which uses AST-guided
perturbations for calibration. AST-PAC improves robustness in larger and more
complex files, but under-mutates small files and is weaker on alphanumeric-rich
code, motivating adaptive and hybrid perturbations for compliance-oriented
auditing.

\newpage

\bibliographystyle{ACM-Reference-Format}
\bibliography{main}

\end{document}